\documentclass[journal]{IEEEtran}
\usepackage[pdftex]{graphicx}
\usepackage{subfigure}
\usepackage{amsmath,amssymb,amsopn,amstext,amsfonts, amsthm}
\usepackage[space]{cite}
\usepackage[colorlinks=false]{hyperref}
\usepackage{url}
\usepackage{enumerate}
\usepackage{xcolor}
\usepackage{colortbl,booktabs}
\usepackage{color,soul}

\hypersetup{hidelinks}
\newtheorem{theorem}{Theorem}

\begin{document}
\title{BALM: Bundle Adjustment for Lidar Mapping}

\author{Zheng Liu and Fu Zhang
\thanks{Z. Liu and F. Zhang are with the Department
of Mechanical Engineering, University of Hong Kong, Hong Kong,
China. {\tt\small u3007335@connect.hku.hk}, {\tt\small fuzhang@hku.hk}}
}


\maketitle
\pagestyle{empty}
\thispagestyle{empty}
\begin{abstract}
A local Bundle Adjustment (BA) on a sliding window of keyframes has been widely used in visual SLAM and proved to be very effective in lowering the drift. But in lidar SLAM, BA method is hardly used because the sparse feature points (e.g., edge and plane) make the exact point matching impossible. In this paper, we formulate the lidar BA as minimizing the distance from a feature point to its matched edge or plane. Unlike the visual SLAM (and prior plane adjustment method in lidar SLAM) where the feature has to be co-determined along with the pose, we show that the feature can be analytically solved and removed from the BA, the resultant BA is only dependent on the scan poses. This greatly reduces the optimization scale and allows large-scale dense plane and edge features to be used. To speedup the optimization, we derive the analytical derivatives  of the cost function, up to second order, in closed form. Moreover, we propose a novel adaptive voxelization method to search feature correspondence efficiently. The proposed formulations are incorporated into a LOAM back-end for map refinement. Results show that, although as a back-end, the local BA can be solved very efficiently, even in real-time at 10Hz when optimizing 20 scans of point-cloud. The local BA also considerably lowers the LOAM drift. Our implementation of the BA optimization and LOAM are open-sourced to benefit the community\footnote{\url{https://github.com/hku-mars/BALM}}.

\end{abstract}

\IEEEpeerreviewmaketitle

\vspace{-0.4cm}

\section{Introduction}

Bundle adjustment (BA) is the problem of jointly solving the 3D structures (i.e., location of feature points) and camera poses \cite{triggs1999bundle}. It has been a fundamental problem in various visual applications, such as structure from motion (SfM) \cite{agarwal2010bundle}, visual SLAM (simultaneous localization and mapping) \cite{mur2015orb}, and visual-inertial navigation \cite{mourikis2007multi,leutenegger2015keyframe}.

Similar bundle adjustment can be defined for lidar mapping where the goal is to jointly determine the lidar pose and the global 3D point-cloud map. This would be a key problem in lowering the drift in lidar SLAM. Constrained by the pairwise matching nature of existing scan registration methods, such as iterative closest points (ICP) \cite{icp1992}, generalized ICP \cite{segal2009generalized}, normal distribution transform \cite{stoyanov2012fast}, and surfel-based registration \cite{behley2018efficient}, commonly used lidar navigation and mapping (LOAM) framework \cite{zhangjiloam} and its variants \cite{legoloam2018, loamlivox} usually build the map by incrementally registering new scans. Such an incremental mapping process would inevitably accumulate registration errors, especially in featureless environments where degeneration occurs \cite{zhang2016degeneracy} or for lidars of small FoV \cite{liu2020low}. One way to lower such drift is performing a local BA over a sliding window of lidar scans, which allows us to re-assess the past scans based on information in new scans. This method has been widely used in visual navigation and proved to be very effective \cite{mourikis2007multi,leutenegger2015keyframe}.

\begin{figure} [t]
	\centering
	\includegraphics{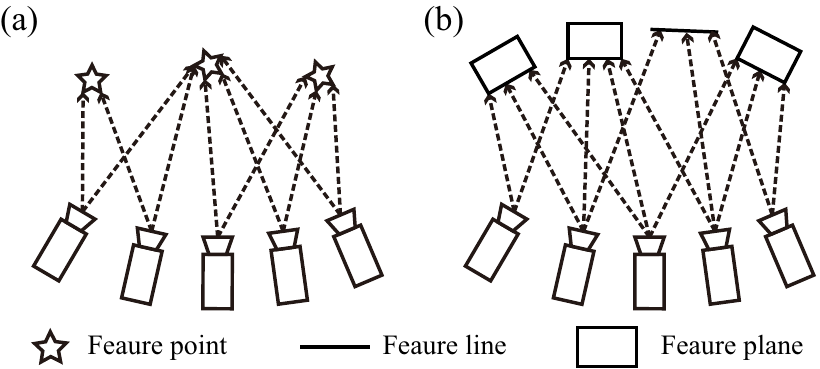}
    \caption{Comparison of BA formulations: (a) visual BA constrains feature points to locate at the same point; (b) our proposed lidar BA constrains feature points to lie on the same edge or plane. }
    \label{figba}
    \vspace{-0.2cm}
\end{figure}

While lidar BA seems simpler than visual BA due to the direct depth measurements, its formulation is actually more complicated. In visual BA, the measurements are high-resolution images where each pixel corresponds to a single feature in the space (see Fig. \ref{figba}(a)). Hence, a natural formulation would be to minimize the difference between the projected feature location and its actual location on the image. However, this natural formulation does not apply to lidar: lidar point-cloud is usually very sparse and even non-repetitive \cite{loamlivox}, making the exact point matching infeasible.

\begin{figure} [t]
	\centering
	\includegraphics{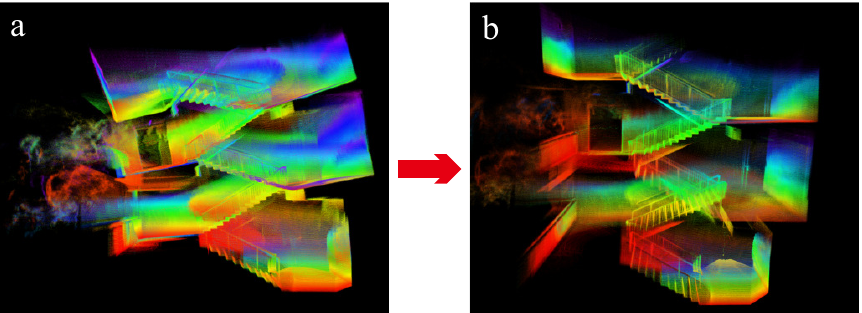}
    \caption{(a) LOAM mapping without map refinement. (b) Refining the map using a local BA on a sliding window of lidar scans. Video available at \url{https://youtu.be/d8R7aJmKifQ}.}
    \label{figback}
    \vspace{-0.5cm}
\end{figure}

In this paper, we propose a formulation of lidar BA and incorporate it into a LOAM framework as the back-end to refine the incrementally built map. More specifically, our contribution is as follows: 1) We formulate the BA on sparse lidar feature points, including both edges and planes, by directly minimizing the distance from the feature point to the edge or plane (see Fig. {\ref{figba}}(b)). Unlike visual BA which simultaneously solves the feature location and camera poses, we show that the feature (edge and plane) parameters in lidar BA can be analytically solved in closed-form solution, leading to a BA optimization over the scan poses only. Eliminating the feature parameters from the BA dramatically reduces the dimension of optimization and hence allows large-scale dense features to be optimized; 2) To enable efficient BA optimization, we analytically derive the gradient and Hessian matrix of the cost function with respect to the scan poses; 3) We propose an adaptive voxelization to search for feature correspondence efficiently; and 4) We incorporate the proposed lidar BA into a LOAM back-end for map refinement and demonstrate its effectiveness on both spinning lidars and lidars of small FoV (e.g., Livox Horizon\footnote{https://www.livoxtech.com/horizon}) by comparing with existing LOAM implementations shown in Fig \ref{figback}. Results show that the local BA effectively lowers the drift. Although it is designed as a back-end, the local BA runs very fast: when optimizing a sliding window of 20 scans, it runs nearly real-time at 10Hz. The BA formulation, optimization libraries, and LOAM implementations are open-sourced to the community.

\vspace{-0.2cm}

\section{Related work}

Our definition of the lidar BA is most similar to the multi-view registration. Early work in this direction \cite{blais1995registering, benjemaa1998solution} directly extend the ICP method \cite{icp1992} to the multi-scans cases, where the cost function is the sum of all distance between two corresponding points in any two scans. Similarly, Neugebauer \cite{neugebauer1997reconstruction} uses the distance between two corresponding surfaces in any two scans. While these methods work well for dense 3D scans (e.g., depth camera), they all require exact point or surface matching that
seldom exists in lidar point-cloud.

The work in \cite{bergevin1996towards, lu1997globally, pulli1999multiview, huber2003fully, borrmann2008globally, grisetti2010tutorial, icpposegraph} register any two scans sharing overlaps using standard pairwise scan registration methods. Then the obtained relative poses are used as measurements to construct a pose graph, from which the poses can be solved by graph optimization. These methods require to perform repeated pairwise scan registration among all scans having overlaps. Moreover, it does not optimize the point-cloud map directly, restricting the attainable level of mapping consistency.

The difficulty of lidar BA (or multi-view registration) lies in defining a metric that effectively evaluates the alignment quality of sparse points from all scans and, in the meantime, allows efficient optimization. The correlation (or entropy)-based scan registration in \cite{tsin2004correlation} naturally extends to multiple scans, however, it requires to compute the correlation between all point pairs, a computation-costly procedure that requires careful engineering \cite{olson2009real} or GPU acceleration \cite{maddern2012lost}.

To lower the computation load, recent work {\cite{kaess2015simultaneous,hsiao2017keyframe,geneva2018lips,zhou2020efficient}} have concentrated on conducting bundle adjustment on plane features only. These work simultaneously optimize the plane parameters and scan poses, leading to an optimization of high-dimension. In contrast, our method considers both planes and edges and analytically solve both features in closed-form before the BA optimization. The resultant BA reduces to minimizing the eigenvalues over the scan poses only. With a cost function (i.e., eigenvalue) similar to {\cite{ferrer2019eigen}} which uses an inefficient gradient descent optimization, we analytically derive the second order derivatives and exploit a highly efficient Gauss-Newton method to speedup the optimization. These two novel contributions, i.e., the elimination of feature (both edge and plane) parameters from the optimization, which significantly reduces the optimization dimension, and the second-order optimization, which significantly speedup the convergence, allows the BA to be conducted in nearly real-time even when optimizing very large number (e.g., a few hundreds) of features. Moreover, unlike prior methods {\cite{kaess2015simultaneous,hsiao2017keyframe,geneva2018lips,zhou2020efficient, ferrer2019eigen}} which typically require to segment planes from raw point-cloud and usually admit true plane features, we propose an adaptive voxelization to match both plane and edge features without a segmentation. This method can further adapt to various environments with both large planes (e.g., ground, wall) and small planar patches (e.g., tree crowns).

There are also some existing work on LOAM with sliding window optimization. Ye {\it et al.} \cite{ye2019tightly} and Shan {\it et al} \cite{shan2020lio} optimize a sliding window of lidar scans by registering each scan in the sliding window to the map built so far. This essentially ignored all concurrent constraints among scans within the sliding window, hence leads to subopotimal solutions. Accounting for all these constraints would lead to repeated pairwise scan registration as in \cite{surmann2003autonomous}. Droeschel {\it et al.} \cite{droeschel2018efficient} uses a multi-resolution occupancy grid map which allows multi-view registration but is too costly in memory or computation \cite{hahnel2003learning}. Compared to these work, our method considers all constraints among all scans either in the sliding window or from the map and can be solved very efficiently.

The rest of the paper is organized as follows: In Section \ref{optimizer}, we present the theoretical framework for BA on sparse lidar points. The adaptive voxelization is presented in Section \ref{ad_voxel}. We present our LOAM implementation with a local BA  in Section \ref{description}. The experiments are detailed in Section \ref{experiment}. Finally, Section \ref{conclude} concludes the paper and presents future work.

\vspace{-0.2cm}

\section{BA formulation and Derivatives} \label{optimizer}

\begin{figure} [t]
	\centering
	\includegraphics[width=8cm]{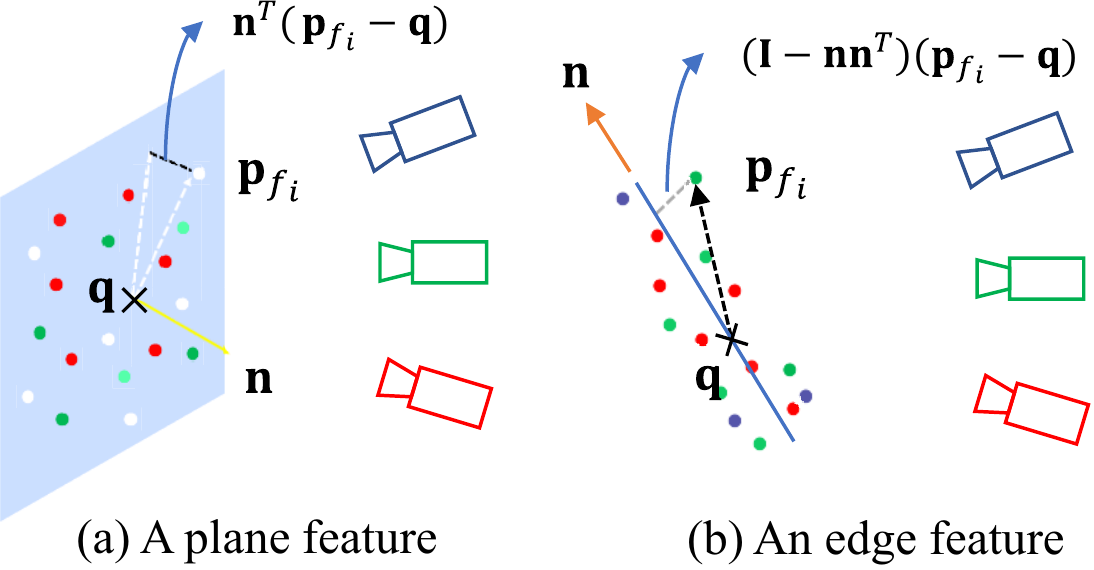}
    \caption{A feature in space and the corresponding feature points drawn from multiple scans: (a) plane feature; (b) edge feature.}
    \label{figmultipleframe}
    \vspace{-0.4cm}
\end{figure}

\subsection{Direct BA formulation}

Given a group of sparse feature points $\mathbf{p}_{f_i} $ $(i=1,\cdots,N)$ drawn from $M$ scans but all correspond to the same feature (plane or edge) (see Fig. \ref{figmultipleframe}). Assume the $i$-th feature point is drawn from the $s_i$-th scan, where $s_i \in \{ 1, \cdots, M\}$, and denote the pose of the $M$ scans as $\mathbf T = \left( \mathbf T_1, \cdots, \mathbf T_M \right)$, where $\mathbf T_j = \left( \mathbf R_j, \mathbf t_j \right) \in SO(3) \times \mathbb{R}^3$ and $j \in \{1, \cdots, M \}$. Then, the feature point in global frame is
\begin{align}
    \label{eqpoint_transformation}
    \mathbf p_i = \mathbf R_{s_i} \mathbf p_{f_i} + \mathbf t_{s_i}; \ i= 1, \cdots, N.
\end{align}

As defined previously, the problem of lidar BA refers to jointly determining the poses of the $M$ scans and the global 3D point-cloud map. Now the 3D map is a single feature (edge or plane), then the BA reduces to jointly determining the poses $\mathbf T$ and location of the single feature, which is represented by a point $\mathbf q$ on the feature and a unit vector $\mathbf n$ ($\mathbf n$ is the normal vector of the plane or the direction of the edge). In case of plane feature, the direct BA formulation is to minimize the summed squared distance from each plane feature point $\mathbf p_i$, which depends on the pose $\mathbf T_{s_i}$, to the plane:
\begin{equation}
    \label{eqplane}
    \begin{aligned}
    & (\mathbf T^*, \mathbf n^*, \mathbf q^*) = \arg \min_{\mathbf T, \mathbf n, \mathbf q}  \frac{1}{N} \sum\nolimits_{i=1}^{N} \left( \mathbf n^T \left( \mathbf p_i - \mathbf q\right) \right)^2 \\
    & \quad = \arg \min_{\mathbf T} \underbrace{ \left( \min_{ \mathbf n, \mathbf q} \frac{1}{N} \sum\nolimits_{i=1}^{N} \left( \mathbf n^T \left( \mathbf p_i - \mathbf q\right) \right)^2  \right)}_{ = \lambda_3 \left( \mathbf A \right); \text{ if } \mathbf n^* = \mathbf u_3, \mathbf q^* = \bar{\mathbf p}},
\end{aligned}
\end{equation}
where $\lambda_k \left( \mathbf A \right)$ denotes the $k$-th largest eigenvalue of matrix $\mathbf A$, $\mathbf u_k$ is the corresponding eigenvector, $\bar{\mathbf p}$ and $\mathbf A$ are:

\begin{align}
    \label{eqcov}
    \bar{\mathbf p} = \frac{1}{N} \sum\nolimits_{i=1}^{N} \mathbf p_i; \ \mathbf A = \frac{1}{N} \sum\nolimits_{i=1}^{N} \left( \mathbf p_i  - \bar{\mathbf p} \right)   \left( \mathbf p_i - \bar{\mathbf p} \right)^T.
\end{align}

Similar to the plane feature, the direct BA formulation for an edge feature is to minimize the summed squared distance from each edge feature point $\mathbf p_i$ to the edge:
\begin{equation}
    \label{eqedge}
    \begin{aligned}
    &(\mathbf T^*, \mathbf n^*, \mathbf q^*) = \arg \min_{\mathbf T, \mathbf n, \mathbf q}  \frac{1}{N} \sum\nolimits_{i=1}^{N} \left \| (\mathbf I - \mathbf n \mathbf n^T) \left( \mathbf p_i - \mathbf q\right) \right \|^2_2 \\
    &= \arg \min_{\mathbf T} \underbrace{ \left( \min_{ \mathbf n, \mathbf q}   \frac{1}{N} \sum\nolimits_{i=1}^{N} \left \| (\mathbf I - \mathbf n \mathbf n^T) \left( \mathbf p_i - \mathbf q\right) \right\|^2_2 \right)}_{ = \text{Tr}(\mathbf A) - \lambda_1 \left( \mathbf A \right) = \lambda_2(\mathbf A) + \lambda_3 (\mathbf A); \text{ if } \mathbf n^* = \mathbf u_1, \mathbf q^* = \bar{\mathbf p}},
\end{aligned}
\end{equation}
where $\text{Tr}(\mathbf A) = \frac{1}{N} \sum_{i=1}^{N} \| \mathbf p_i - \bar{\mathbf p} \|_2^2 $ denotes the trace of $\mathbf A$.

Note that in ({\ref{eqplane}}) and ({\ref{eqedge}}), the optimal point $\mathbf q^*$ is not unique, as the point is free to move within the plane (or along the edge). However, this has no effect on the resultant cost function to be optimized. Furthermore, ({\ref{eqplane}}) and ({\ref{eqedge}}) imply that the optimal feature (plane or edge) parameter can be analytically obtained before the BA, and the resultant BA problem is only dependent on the poses $\mathbf T$. This agrees well to our intuition that the 3D point-cloud map (hence the plane or edge features) are determined once the scan poses are known. Moreover, the optimization on the poses $\mathbf T$ reduces to minimizing the eigenvalues of the matrix $\mathbf A$ in ({\ref{eqcov}}). i.e., the BA leads to minimizing

\begin{equation}
    \label{eqcost}
    \lambda_k(\mathbf p (\mathbf T)),
\end{equation}
over $\mathbf T$, where $\mathbf{p}=[\mathbf{p}_1^T \cdots \mathbf{p}_N^T]^T$ is the vector of all feature points corresponding to the same feature.

To allow efficient optimization with the cost in (\ref{eqcost}), we analytically derive the closed-form derivatives, up to second order, with respect to the pose $\mathbf T$. Due to the chain rule, we derive the derivatives with respect to the point vector $\mathbf p$ first.
\vspace{-0.4cm}
\subsection{The Derivatives}

\begin{theorem}\label{theorem1}
For a group of points, $\mathbf{p}_i \ (i=1,\cdots,N)$ and the covariance matrix $\mathbf{A}$ defined in (\ref{eqcov}). Assume $\mathbf{A}$ has eigenvalues $\lambda_k $ corresponding to eigenvectors $\mathbf{u}_k\ (k=1,2,3)$, then
\begin{align}\label{eqDer}
    \frac{\partial \lambda_k}{\partial \mathbf{p}_i} = \frac{2}{N}(\mathbf{p}_i-\bar{\mathbf{p}})^T \mathbf{u}_k \mathbf{u}_k^T,
\end{align}
where the $\bar{\mathbf{p}}$ is the average of the $N$ points as in (\ref{eqcov}).
\end{theorem}

\begin{theorem}\label{theorem2}
For a group of points, $\mathbf{p}_i\ (i=1,\cdots,N)$ and the covariance matrix $\mathbf{A}$ defined in (\ref{eqcov}). Assume $\mathbf{A}$ has eigenvalues $\lambda_k $ corresponding to eigenvectors $\mathbf{u}_k\ (k=1,2,3)$. Moreover, $\lambda_i \neq \lambda_k$ when $i \neq k$, then

\begin{align} \label{th7}
    \frac{\partial^2 \lambda_k}{\partial\mathbf{p}_j \partial \mathbf{p}_i} = \left\{
    \begin{aligned}
    \frac{2}{N}\bigg(\frac{N-1}{N}\mathbf{u}_k \mathbf{u}_k^T  + \mathbf{u}_k(\mathbf{p}_i-\bar{\mathbf{p}})^T \mathbf{UF}^{\mathbf{p}_j}_{k} \\
    + \mathbf{UF}^{\mathbf{p}_j}_{k} \Big(\mathbf{u}_k^T (\mathbf{p}_i-\bar{\mathbf{p}})\Big)\bigg), \quad i=j \\
    \frac{2}{N}\bigg(-\frac{1}{N}\mathbf{u}_k \mathbf{u}_k^T + \mathbf{u}_k(\mathbf{p}_i-\bar{\mathbf{p}})^T \mathbf{UF}^{\mathbf{p}_j}_{k} \\
    + \mathbf{UF}^{\mathbf{p}_j}_{k} \Big(\mathbf{u}_k^T (\mathbf{p}_i-\bar{\mathbf{p}})\Big)\bigg), \quad i\neq j
    \end{aligned}
    \right.
\end{align}

\begin{align*}
    \mathbf{F}^{\mathbf{p}_j}_{k} = \begin{bmatrix}
    \mathbf{F}^{\mathbf{p}_j}_{1,k} \\
    \mathbf{F}^{\mathbf{p}_j}_{2,k} \\
    \mathbf{F}^{\mathbf{p}_j}_{3,k}
    \end{bmatrix} \in \mathbb{R}^{3 \times 3},  \quad
    \mathbf{U} = \begin{bmatrix}
    \mathbf u_1 & \mathbf u_2 & \mathbf u_3
    \end{bmatrix},
\end{align*}

\begin{align*}
    \mathbf{F}^{\mathbf{p}_j}_{m,n} &=\left\{\begin{aligned}
     \frac{(\mathbf{p}_j-\bar{\mathbf{p}})^T}{N(\lambda_n-\lambda_m)} (\mathbf{u}_m \mathbf{u}_n^T + \mathbf{u}_n \mathbf{u}_m^T), m\neq n \\
    \mathbf{0}_{1 \times 3}  \qquad\qquad\qquad, m=n
    \end{aligned}\right.
\end{align*}
\end{theorem}

\subsection{Second order approximation}

With the first and second order derivatives in previous sections, we can approximate the cost function (\ref{eqcost}) by its second order approximation as below:

\vspace{-0.3cm}

\begin{align}
    \label{eqSecondOrder}
    \lambda_k(\mathbf{p}+\delta\mathbf{p}) \approx \lambda_k(\mathbf{p}) + &\mathbf{J}(\mathbf{p})\delta \mathbf{p} + \frac{1}{2}\delta \mathbf{p}^T\mathbf{H}(\mathbf{p})\delta\mathbf{p},
\end{align}
where $\mathbf{J}(\mathbf{p})$ is the Jacobian matrix with $i$-th elements in (\ref{eqDer}) and $\mathbf{H}(\mathbf{p})$ is the Hessian matrix with $i$-th row, $j$-th column elements in (\ref{th7}).

Recall that the point vector $\mathbf p$ is further dependent on the scan poses $\mathbf T$ as in (\ref{eqpoint_transformation}). Perturbing a pose $\mathbf T_j$ in its tangent plane $\delta \mathbf T_j = \begin{bmatrix}
\boldsymbol{\phi}_j^T & \delta \mathbf t_j^T
\end{bmatrix}^T$ using the $\boxplus$ operation defined in \cite{hertzberg2013integrating}, we have
\begin{align}
    \mathbf T_j = (\mathbf R_j, \mathbf t_j); \ \mathbf T_j \boxplus \delta \mathbf T_j = (\mathbf R_j \exp \left( \boldsymbol{\phi}_j^{\land} \right), \mathbf t_j +  \delta \mathbf t_j )
\end{align}
and
\begin{align}
    &\mathbf{p}_{i} \! = \! \mathbf R_{s_i} \exp(\boldsymbol{\phi}_{s_i}^{\land})\mathbf{p}_{f_i} \! + \! \mathbf{t}_{s_i}; \ \frac{\delta\mathbf{p}_{i}}{\delta\mathbf{T}_{s_i}} \! = \!
    \begin{bmatrix}
    -\mathbf{R}_{s_i} (\mathbf{p}_{f_i})^{\land} & \mathbf{I}
    \end{bmatrix}  \\
    & \mathbf{D} =\frac{\delta\mathbf{p}}{\delta\mathbf{T}} = \begin{bmatrix}
     & \vdots & \\
     \cdots & \mathbf D_{ij} & \cdots \\
     & \vdots &
    \end{bmatrix}\in \mathbb{R}^{3N \times 6M} \\
    \label{eqD}
    & \mathbf D_{ij} = \left\{ \begin{array}{rl} \frac{\delta\mathbf{p}_{i}}{\delta\mathbf{T}_{s_i}} & \mbox{for } j = s_i \in \{ 1, \cdots, M\} \\
    \mathbf 0_{3 \times 6} & \mbox{for else}
\end{array}\right.
\end{align}

Substituting (\ref{eqD}) into (\ref{eqSecondOrder}) leads to
\begin{align}
    \label{eqSecondApprox}
    \lambda_k(\mathbf{T}\boxplus\delta\mathbf{T}) \approx \lambda_k(\mathbf{T}) + \underbrace{\mathbf{JD}}_{\bar{\mathbf{J}}}\delta\mathbf{T} + \frac{1}{2}\delta\mathbf{T}^T \underbrace{\mathbf{D}^T \mathbf{HD}}_{\bar{\mathbf{H}}}\delta \mathbf{T}
\end{align}

Finally, we use a Levenberg-Marquardt (LM) \cite{lm1978} method to minimize the cost $\lambda_k$ by repeatedly approximating it by the second order approximation (\ref{eqSecondApprox}). In each iteration, the solution is solved from
\vspace{-0.1cm}
\begin{align}
    (\bar{\mathbf{H}}(\mathbf T)+\mu \mathbf{I})\delta \mathbf{T}^* &=-\bar{\mathbf{J}}(\mathbf{T})^T, \label{lm2}
\end{align}
where $\mu$ is the stepsize determined from the LM method.

\section{Adaptive Voxelization} \label{ad_voxel}
\begin{figure} [t]
	\centering
	\includegraphics[width=8cm]{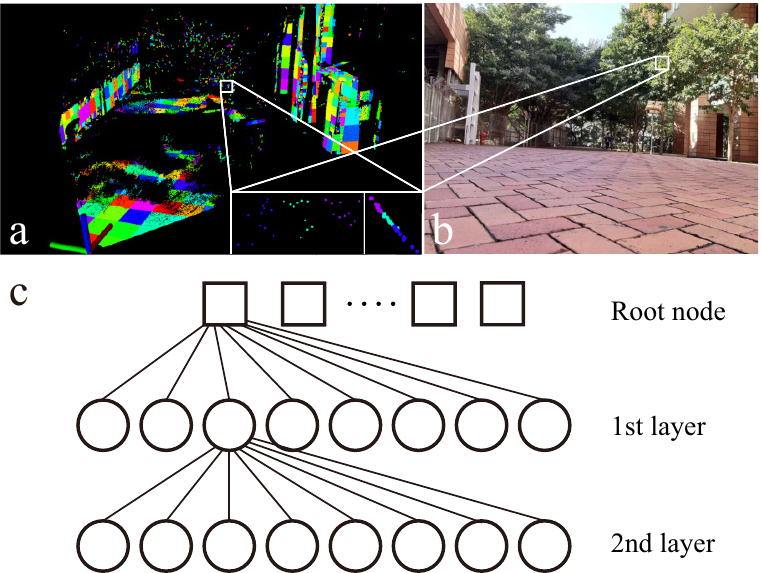}
    \caption{\label{figvoxel} (a) An exemplary voxel map, different color represents different voxels. Pictures in the lower right white box are the zoomed view of plane points on the tree crown, which contains 3 voxels with size 0.125m (left: front view; right: side view). (b) The actual environment photo. (c) All octrees are indexed in a Hash table.}
    \vspace{-0.4cm}
\end{figure}

The BA formulation in Section. \ref{optimizer} requires to find all feature points corresponding to the same feature (edge or plane). To do so, we propose a novel adaptive voxelization method: assume that a rough initial pose of different scans are available (e.g., from a LOAM odometry), we repeatedly voxelize the 3D space from a default size (e.g., $1m$): if all feature points (from all scans) in the current voxel lie on a plane or edge (e.g., by examining the eigenvalue of the point covariance matrix (\ref{eqcov})), the current voxel is kept in memory along with the contained feature points; otherwise, the current voxel breaks into eight octants and proceeds to check each octant until reaching the minimal size (e.g., $0.125m$). The proposed adaptive voxelization generates a voxel map, where different voxels may have different size adapted to the environment. For each voxel, it corresponds to one feature, and hence one cost item as in (\ref{eqSecondApprox}). An exemplary voxel map is seen in Fig. \ref{figvoxel}(a).

The adaptive voxelization has many advantages: 1) It is naturally compatible with existing data structures such as octrees, hence its implementation and efficiency can be greatly facilitated; 2) It is usually more efficient than constructing a full Kd-tree on feature points \cite{zhangjiloam} as early termination may occur when the contained feature points lie on the same plane or edge. Such an advantage will be more obvious when the environment has large planes or long edges; 3) A map with adaptive voxels will lower the time for searching feature correspondences in lidar odometry. It is only necessary to search the voxel a feature point lies in or near to, instead of the nearest points that require more exhaustive search \cite{zhangjiloam}.

In our implementation, we construct two voxel maps, one for edge features and one for planar features. The voxel map, by its construction, naturally suits to an octree structure. To reduce the depth of the octree, we use a set of octrees indexed by a Hash table (see Fig. \ref{figvoxel}(c)). Each octree corresponds a non-empty cube of the default voxel size (e.g., $1m$) in the space. Different octrees may have different depth, depending on the geometry of that cube in the space. Each leaf node (i.e., a voxel) in an octree saves feature points all corresponding to the same feature (e.g., plane or edge).

{\it Remark 1. } If a voxel contains too many points, the Hessian matrix in (\ref{eqSecondOrder}) would have a very high dimension, in this case, we could average the points from the same scan. The averaged points have fewer number and lie on the same plane determined by the raw feature points. This allows to save much computation without degrading the mapping consistency.

{\it Remark 2. } The Hessian matrix computed in Theorem \ref{theorem2} requires $\lambda_i \neq \lambda_k$ when $i \neq k$. For a voxel whose $\lambda_k$ has algebraic multiplicity more than one, we simply skip it.

{\it Remark 3. } Although we keep saying edge features and plane features, the method naturally extends to non-planar features (e.g., curved surfaces) by constructing the voxel map at a finer level and allowing larger variance when examining whether the contained points lie on the same plane.

{\it Remark 4. } Two conditions are set to stop the recursive sub-division: one is the maximal depth of the tree and the other is the minimum number of points in a voxel.

\section{LOAM with Local BA} \label{description}
\begin{figure} [htbp]
	\centering
	\includegraphics[width=8.5cm]{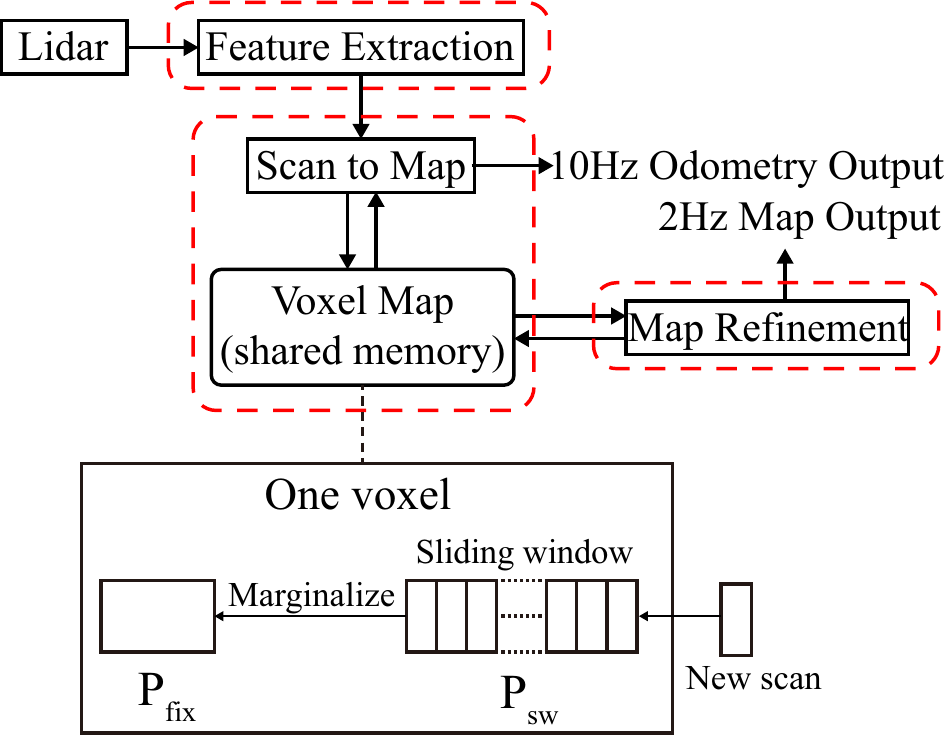}
    \caption{\label{figstruct} Overview of LOAM with local BA.}
    \vspace{-0.2cm}
\end{figure}

In this section, we incorporate the proposed BA formulation and its optimization methods into a LOAM framework. The system overview is shown in Fig. \ref{figstruct}. It consists of three parallel threads: feature extraction, odometry, and map-refinement. The feature extraction thread extracts the edge and plane features similar to \cite{zhangjiloam} and \cite{loamlivox}.

Once receiving a new scan of feature points, the odometry estimates the lidar pose by registering the new scan to the existing map. Unlike the existing methods \cite{zhangjiloam, loamlivox} where each feature point is matched to some nearest points in the map, we leverage the adaptive voxel map to speedup the matching process. More specifically, when constructing the voxel map, we compute the center point and normal (or direction) vector of the plane (or edge) in a voxel. Then for a point in the new scan, we search the nearest voxel (represented by its center point) by computing the distance between the point and the plane or edge feature in the voxel.

With the odometry, the new scan can be roughly registered to the global frame and be pushed to the voxel map: for each point in the new scan, search the voxel it lies in and add this point to the leaf node of the corresponding octree. If no voxel is found in the existing map for the point in the new scan, create a new octree, index its root in the Hash table, and add this point to the root node. After all feature points of the new scan are distributed to the leaf node of existing octrees or the root node of newly created octrees, we update the voxel map as the way it is constructed: if points in a node (leaf or node) do not make a single feature (plane or edge), divide the node into eight and check each of them.

After pushing a certain number of new scans to the voxel map, a map-refinement is triggered. The map-refinement performs a local BA on a sliding window of lidar poses. Any voxel containing points within the sliding window (i.e., $\mathbf{P}_{sw}$) are used to construct cost items as ({\ref{eqplane}}) or ({\ref{eqedge}}). Then, the map-refinement repeatedly minimizes the second order approximation (\ref{eqSecondApprox}) of the total cost consisting of all relevant voxels. This refines all the lidar poses within the sliding window. The updated poses are then used to update the center points and normal vectors of all involved voxels.

Once the sliding window is full, points from older scans are merged to the map points $\mathbf P_{fix}$. A nice property of the point covariance matrix (\ref{eqcov}) is the existence of recursive form \cite{lin2019fast}, allowing all points outside the sliding window to be summarized in a few compact matrices and vectors without saving the raw points (see lower part of Fig. \ref{figstruct}). The merged points $\mathbf P_{fix}$ will be retained in the voxel map for odometry and map-refinements.
\vspace{-0.3cm}

\section{Experiments} \label{experiment}
We present experimental results to verify the effectiveness of the proposed BA in LOAM. In the experiment, the lidar odometry runs at $10$Hz, the map-refinement is triggered after receiving 5 scans hence running at $2$Hz. We use a sliding window of $20$ most recent scans. All the experiments run on a laptop computer with CPU i7-10750H and 16Gib memory. More experiment details can be found in the video available at \url{https://youtu.be/d8R7aJmKifQ}.
\vspace{-0.4cm}

\subsection{Livox Horizon}

\begin{figure} [htbp]
	\centering
	\includegraphics[width=8.5cm]{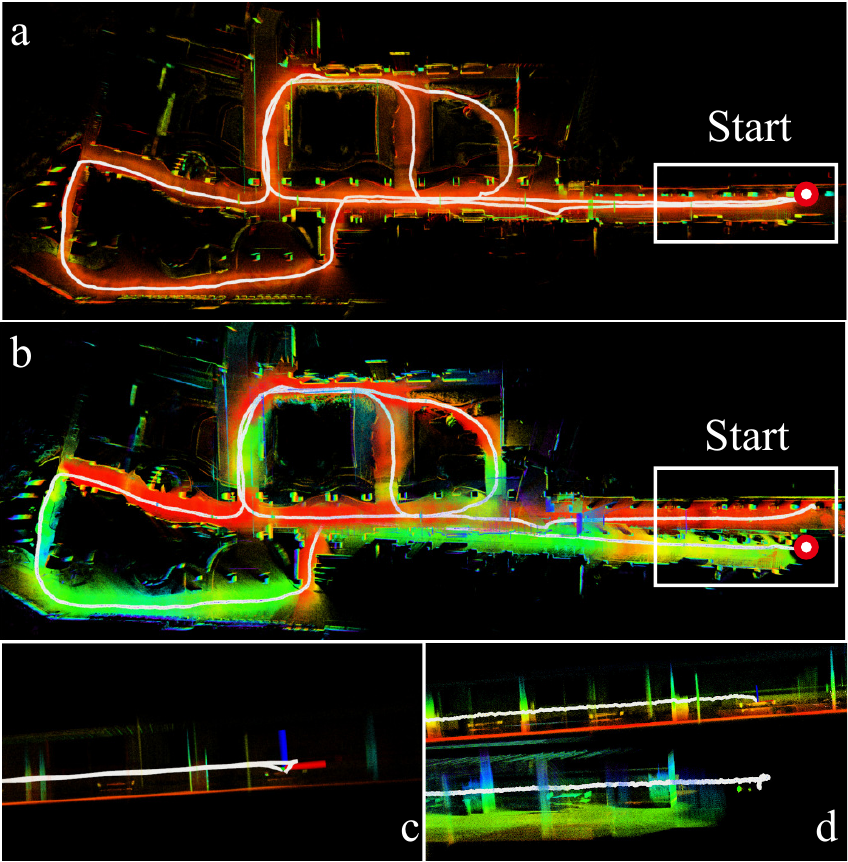}
    \caption{Results of outdoor walking dataset: (a) the overall map built by BALM, (b) the map built by LOAM. (c) and (d) show the side view of map and odometry near to the start/end point (the white box in (a) and (b)) of our method BALM and LOAM, respectively.}
    \label{hku_park}
    \vspace{-0.2cm}
\end{figure}

We test our algorithm on Livox Horizon lidar, which has a $25^{\circ} \times 82^{\circ}$ FoV, and compare its performance with that of a state-of-the-art implementation of LOAM \cite{zhangjiloam} for this lidar\footnote{\url{https://github.com/Livox-SDK/livox_mapping} \label{livoxmapping}}. The lidar in this experiment is handheld and moving in HKU campus (Fig. \ref{hku_park}). The total path length is about 817m. We return to the start position after 20 minutes of walking. Fig. \ref{hku_park} (a) and (b) show the odometry and mapping results of BALM and LOAM, respectively. It is seen that our method successfully returns to the start point while LOAM leads to significant drift. The elevation error of our method is also much smaller (e.g., 0.27m versus 3.98m, see Fig. \ref{hku_park} (c) and (d)). The total translation error is summarized in Table \ref{park_hku}.

We additionally conducted an indoor experiment where the sensor is handheld and moving along a stairway. Results in Fig. \ref{figback} validates the effectiveness of the proposed local BA.

\begin{table}
    \caption{Drift comparison on Livox Horizon lidar data.}
    \centering
    \begin{tabular}{cccc}
    \toprule
    Distance (817m) & LOAM (m) & BALM (m) \\
    \midrule
    Translation error & 6.228 (0.762\%) & 0.31 (0.038\%) \\
    \bottomrule
    \end{tabular}
    \label{park_hku}
    \vspace{-0.4cm}
\end{table}
\begin{figure} [t]
	\centering
	\includegraphics[width=8.5cm]{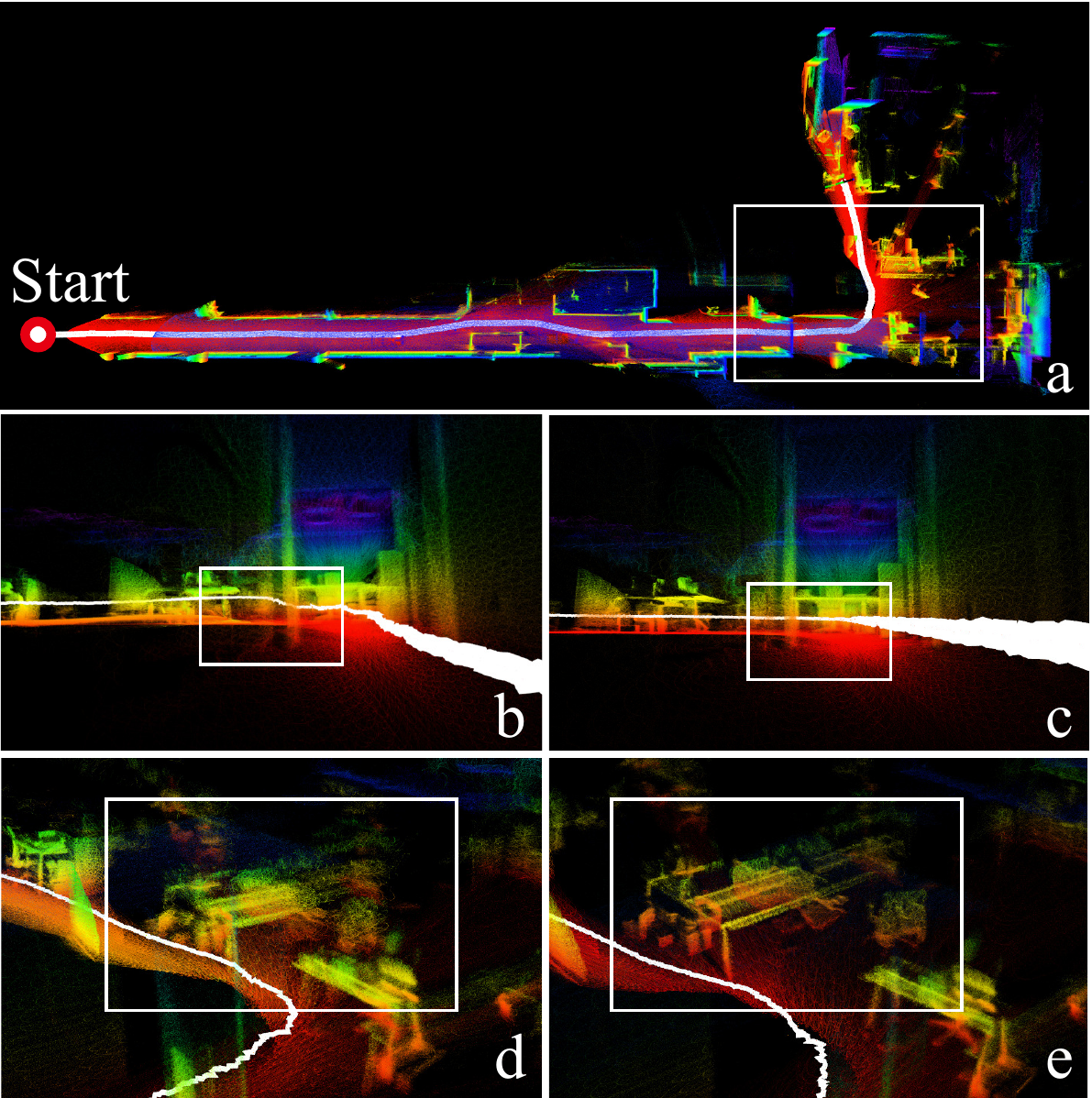}
    \caption{Indoor mapping and odometry results: (a) the overall scene; (b) and (d) show the odometry and mapping results of LOAM; (c) and (e) show our method BLAM. }
    \label{indoor}
\end{figure}

\vspace{-0.3cm}

\subsection{Livox MID-40}

In this experiment, we test our algorithm on Livox Mid-40 lidars\footnote{\url{https://www.livoxtech.com/mid-40-and-mid-100}} mounted on a UGV and compare its performance with LOAM implementation\textsuperscript{\ref{livoxmapping}}. For a classic $360^{\circ}$ spinning lidar, it is easy to turn in a corridor corner.  But for the Livox Mid-40 lidar with a small $40^{\circ}$ FoV,  degeneration occurs easily because of the lessen feature points. This leads to a zigzag pose trajectory in LOAM (Fig. \ref{indoor}(d)). The degraded pose estimation causes map inconsistencies, which in turn worsen the following pose estimation furthermore. As a result, the LOAM odometry is falsely ``raised" at the corner (Fig. \ref{indoor}(b)). On the other hand, the BALM, although produces a similarly zigzag pose trajectory due to degeneration in the front-end (i.e., scan to map) similar to LOAM, has a much consistent map (Fig. \ref{indoor}(e)), which in turn significantly lowers the drift as a result of the local BA (Fig. \ref{indoor}(c)).

\begin{figure} [t]
	\centering
	\includegraphics[width=7cm]{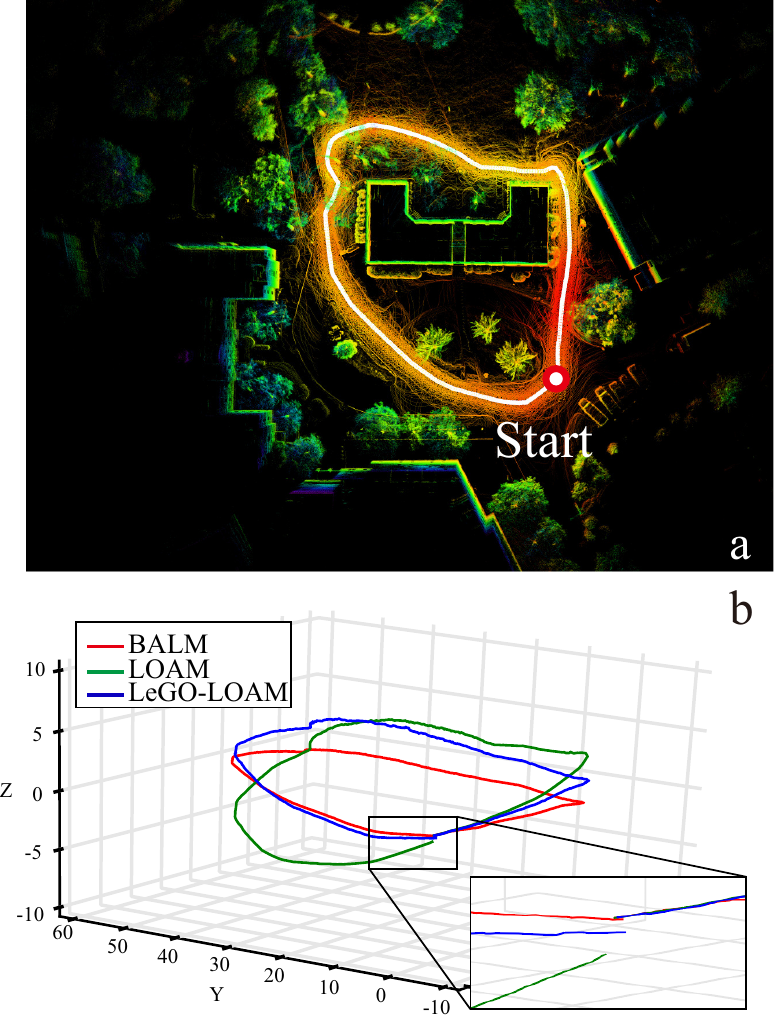}
    \caption{Outdoor mapping and odometry results. (a) The overview of the scene; (b) The paths of LOAM, LeGO-LOAM and BALM}
    \label{outdoor}
    \vspace{-0.4cm}
\end{figure}

\vspace{-0.2cm}

\subsection{Velodyne VLP-16}
We further test our algorithm on Velodyne VLP-16 lidar. We use the data offered by LeGO-LOAM \cite{legoloam2018} available on Github\footnote{\url{https://github.com/RobustFieldAutonomyLab/LeGO-LOAM}}, and perform comparison study. The path has the same starting and end point. The scene can be seen in Fig. \ref{outdoor}(a) and the path is colored in white. The paths of LOAM, LeGO-LOAM and BALM are shown in Fig.\ref{outdoor}(b). The drift when returning to start is given in Table \ref{legodata}.

\begin{table}
    \caption{Drift comparison on Velodyne-16 lidar data.}
    \centering
    \begin{tabular}{cccc}
    \toprule
    Distance(210m) & LOAM (cm) & LeGO-LOAM (cm) & BALM (cm) \\
    \midrule
    Translation error & 56.8 (0.27\%) & 38.5 (0.18\%) & 28.0 (0.13\%)\\
    \bottomrule
    \end{tabular}
    \label{legodata}
\end{table}

\subsection{Running time}
\begin{figure} [t]
	\centering
	\includegraphics[width=8.5cm]{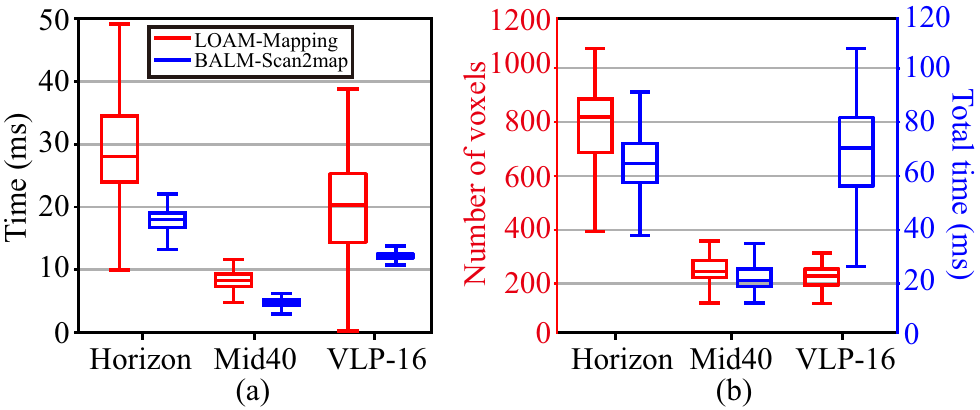}
    \caption{(a) Time for scan to map alignment in LOAM and our method; (b) The number of voxels and computation time for a local BA over 20 scans. }
    \label{time_statistics}
    \vspace{-0.4cm}
\end{figure}

In LOAM, a feature point in new scan should find five closest points, but in BALM, the feature point just need to find the closest voxel (plane or edge), which can reduce the searching time in the scan to map. The comparison is shown in Fig. \ref{time_statistics}(a) where the running time is for building kd-tree, finding closed points/voxels and LM optimization. The data of running time is obtained by experiment A, B and C. To make a fair comparison, a fixed two-step LM optimization is used for both methods.

Finally, Fig. \ref{time_statistics}(b)  shows the number of voxels for a sliding window of $20$ most recent scans and the time for local BA and voxel map update. It is seen that in most cases, the local BA and voxel map update can complete in $100ms$. This implies that the local BA can nearly run in real-time as the odometry (i.e., $10Hz$).

\vspace{-0.4cm}

\section{Conclusion and Future Works} \label{conclude}
This paper formulated a framework for lidar bundle adjustment (BA) and developed theoretical derivatives allowing efficient optimization. A novel adaptive voxelization is proposed to support the lidar BA. Then the proposed BA and optimization methods are further incorporated into a LOAM framework to serve as the back-end for map refinement. Experiments on various lidars and environments validate the effectiveness of the proposed methods.

The current implementation of local BA in LOAM uses a sliding window of temporal scans, leading to redundant information in adjacent scans sharing large overlaps. Moreover, a drawback of our voxelization is the requirement of good initial poses alignment. However, the current lidar odometry uses a simple scan-to-map front-end without compensating any motion distortion or leveraging any motion model. Future works will adopt keyframes in a local sliding window and further incorporate motion models. Besides the LOAM, the proposed BA can also be used for global mapping and extrinsic calibration, which will also be explored in the future.

\vspace{-0.4cm}

\appendices
\section{}
\subsection{Proof of theorem \ref{theorem1}}
Denote a point $\mathbf{p}_i = \begin{bmatrix} x_i & y_i & z_i \end{bmatrix}^T $ and the eigenvector matrix $\mathbf{U} = \begin{bmatrix} \mathbf u_1 & \mathbf u_2 & \mathbf u_3 \end{bmatrix}^T $. Further denote $p$ an element of $\mathbf p_i$, $p$ is one of $x_i, y_i$ and $z_i$. Then by definition, we have
\begin{align}
    \mathbf{\Lambda} &= \mathbf{U}^T \mathbf{A}\mathbf{U} \\
    \frac{\partial\mathbf{\Lambda}}{\partial p} &= \left( \frac{\partial\mathbf{U}}{\partial p}  \right)^T \mathbf A \mathbf U + \mathbf U^T  \frac{\partial\mathbf{A}}{\partial p} \mathbf U + \mathbf U^T \mathbf A  \frac{\partial\mathbf{U}}{\partial p} \label{th2} \\
    \mathbf{U}^{T}\mathbf{A} &= {\mathbf{\Lambda}}\mathbf{U}^T; \ \mathbf A \mathbf U = \mathbf U \mathbf A \label{th3}
\end{align}

Plugging (\ref{th3}) into (\ref{th2}) yields:

\begin{align}
    \frac{\partial\mathbf{\Lambda}}{\partial p} =\mathbf{U}^{T}\frac{\partial\mathbf{A}}{\partial p}\mathbf{U} +  {\mathbf{\Lambda}} \underbrace{\mathbf{U}^{T} \frac{\partial\mathbf{U}}{\partial p}}_{\mathbf C^p} +  \underbrace{\left( \frac{\partial\mathbf{U}}{\partial p}  \right)^T  \mathbf U}_{(\mathbf{C}^p)^T} \mathbf \Lambda \label{th4}
\end{align}

As $\mathbf{U}^{T}\mathbf{U}=\mathbf{I}$, where $\mathbf{I}$ is the identity matrix, partial differentiating both sides with respect to $p$ leads to
\begin{align*}
    \mathbf{U}^{T}\frac{\partial\mathbf{U}}{\partial p} + \Big(\frac{\partial\mathbf{U}}{\partial p}\Big)^T\mathbf{U}= \mathbf 0 \implies \mathbf C^p + (\mathbf{C}^p)^T = \mathbf 0.
\end{align*}

It is seen that $\mathbf C^p $ is a skew symmetric matrix whose diagonal elements are zeros. Moreover, since $\mathbf{\Lambda}$ is diagonal, the last two items of the right side of (\ref{th4}) sum to zero on diagonal positions. Only considering the diagonal elements in (\ref{th4}) leads to
\begin{align*}
    \frac{\partial \lambda_k}{\partial p} = \mathbf{u}_k^T \frac{\partial\mathbf{A}}{\partial p} \mathbf{u}_k = \frac{\partial \mathbf{u}_k^T \mathbf{A} \mathbf{u}_k}{\partial p} \quad (k=1,2,3)
\end{align*}
where in the second equation the vector $\mathbf{u}_k$ is viewed constant. Stacking the partial differentiation of $\lambda_k$ with respect to all elements of $\mathbf p_i$ leads to
\begin{align*}
    \frac{\partial \lambda_k}{\partial \mathbf{p}_i} &= \Big[
        \frac{\partial \mathbf{u}_k^T \mathbf{A} \mathbf{u}_k}{\partial x_i} \quad \frac{\partial \mathbf{u}_k^T \mathbf{A} \mathbf{u}_k}{\partial y_i} \quad \frac{\partial \mathbf{u}_k^T \mathbf{A} \mathbf{u}_k}{\partial z_i} \Big] = \frac{\partial \mathbf{u}_k^T \mathbf{A} \mathbf{u}_k}{\partial \mathbf{p}_i}
\end{align*}

Recall the definition of matrix $\mathbf{A}$ in (\ref{eqcov}) and that
\begin{align*}
    \frac{\partial\mathbf{p}_j}{\partial\mathbf{p}_i} = \mathbf{I} ,(i=j) &\qquad
    \frac{\partial\mathbf{p}_j}{\partial\mathbf{p}_i}=\mathbf{0} , (i\neq j),
\end{align*}
Then, we can obtain
\begin{align}
    \frac{\partial \lambda_k}{\partial \mathbf{p}_i} &= \frac{1}{N} \sum_{j=1}^N  \frac{\partial \mathbf{u}_k^T(\mathbf{p}_j-\bar{\mathbf{p}})(\mathbf{p}_j-\bar{\mathbf{p}})^T \mathbf{u}_k}{\partial\mathbf{p}_i} \notag \\
    &= \frac{2}{N} \sum_{j=1}^N (\mathbf{p}_j-\bar{\mathbf{p}})^T \mathbf{u}_k\frac{\partial \mathbf{u}_k^T(\mathbf{p}_j-\bar{\mathbf{p}})}{\partial\mathbf{p}_i} \notag \\
    &= \frac{2}{N}(\mathbf{p}_i-\bar{\mathbf{p}})^T \mathbf{u}_k \mathbf{u}_k^T(\mathbf{I}-\frac{1}{N}\mathbf{I}) \notag \\ &+\frac{2}{N}\sum\limits_{j=1,j\neq i}^N(\mathbf{p}_j-\bar{\mathbf{p}})^T \mathbf{u}_k \mathbf{u}_k^T(-\frac{1}{N}\mathbf{I}) \notag \\
    &= \frac{2}{N}(\mathbf{p}_i-\bar{\mathbf{p}})^T \mathbf{u}_k \mathbf{u}_k^T. \quad  \blacksquare \label{th9}
\end{align}

\subsection{Proof of theorem \ref{theorem2}}
Consider two points, $\mathbf{p}_i= [x_i \quad y_i \quad z_i] ^T$ and $\mathbf{p}_j= [x_j \quad y_j \quad z_j] ^T$. Denote $q$ a element of $\mathbf p_j$, $q$ is one of $x_j$, $y_j$ and $z_j$. Since the eigenvector matrix $\mathbf{U}$ is orthogonal, so
\begin{align*}
    \mathbf{U}^{T}\frac{\partial\mathbf{U}}{\partial q} + \Big(\frac{\partial\mathbf{U}}{\partial q}\Big)^T\mathbf{U}=0
\end{align*}

Define
\begin{align*}
    \mathbf{C}^{q} = \mathbf{U}^{T}\frac{\partial\mathbf{U}}{\partial q}, \quad \mathbf{C}^{q} + (\mathbf{C}^{q})^T = 0
\end{align*}

The elements on the diagonal of $\mathbf{C}^{q}$ is zero. Similarly with (\ref{th4}) and replace $p$ with $q$
\begin{align}\label{e:eqn}
    \frac{\partial\mathbf{\Lambda}}{\partial q} = &\mathbf{U}^T\frac{\partial\mathbf{A}}{\partial q}\mathbf{U} + \mathbf{\Lambda C}^{q} - \mathbf{C}^{q}\mathbf{\Lambda}
\end{align}

Since $\mathbf \Lambda$ is diagonal and hence $\frac{\partial\mathbf{\Lambda}}{\partial q}$, for off-diagonal elements in (\ref{e:eqn}), we have
\begin{align*}
    0 = &\mathbf{u}_m^T  \frac{\partial\mathbf{A}}{\partial q}\mathbf{u}_n + \lambda_m \mathbf{C}^{q}_{m,n} - \mathbf{C}^{q}_{m,n}\lambda_n
\end{align*}
$\mathbf{C}^{q}_{m,n}$ is the $m$-th row and $n$-th column element in $\mathbf{C}^{q}$ as below if $\lambda_m \neq \lambda_n $
\begin{align} \label{th5}
    \mathbf{C}^{q}_{m,n} &=\left\{\begin{aligned}
     \frac{1}{\lambda_n - \lambda_m} \mathbf{u}_m^T \frac{\partial \mathbf{A} }{\partial  q} \mathbf{u}_n , m\neq n \\
    0  \qquad\qquad, m=n
    \end{aligned}\right.
\end{align}

According the definition of $\mathbf{C}^{q}$,
\begin{align*}
    \frac{\partial\mathbf{u}_k}{\partial q} = \frac{\partial\mathbf{U} \mathbf e_k }{\partial q} = \mathbf{U} \mathbf{C}^{q} \mathbf{e}_k
\end{align*}
where $\mathbf{e}_k$ is a $3 \times 1$ vector in which the $k$-th element is $1$ and the rests $0$. Stacking the partial differentiation of $\mathbf{u}_k$ with respect to all elements of $\mathbf{p}_j$ leads to
\begin{align}
    \frac{\partial\mathbf{u}_k}{\partial \mathbf{p}_j} &= \Big[\frac{\partial\mathbf{U}  \mathbf e_k }{\partial x_j} \quad \frac{\partial\mathbf{U}  \mathbf e_k }{\partial y_j} \quad \frac{\partial\mathbf{U}  \mathbf e_k}{\partial z_j} \Big] \notag \\
    &= \big[\mathbf{U} \mathbf{C}^{x_j} \mathbf{e}_k \quad \mathbf{U} \mathbf{C}^{y_j} \mathbf{e}_k \quad \mathbf{U} \mathbf{C}^{z_j} \mathbf{e}_k \big] \notag \\
    &= \mathbf{U} \big[\mathbf{C}^{x_j} \mathbf{e}_k \quad \mathbf{C}^{y_j} \mathbf{e}_k \quad  \mathbf{C}^{z_j} \mathbf{e}_k \big] \notag \\
    &= \mathbf{U} \begin{bmatrix}
    \mathbf{C}^{x_j}_{1,k} & \mathbf{C}^{y_j}_{1,k} & \mathbf{C}^{z_j}_{1,k} \\
    \mathbf{C}^{x_j}_{2,k} & \mathbf{C}^{y_j}_{2,k} & \mathbf{C}^{z_j}_{2,k} \\
    \mathbf{C}^{x_j}_{3,k} & \mathbf{C}^{y_j}_{3,k} & \mathbf{C}^{z_j}_{3,k} \label{th6}
    \end{bmatrix}
\end{align}

Define
\begin{align*}
    \mathbf{F}_{m,n}^{\mathbf{p}_j} = \begin{bmatrix}
    \mathbf{C}^{x_j}_{m,n} & \mathbf{C}^{y_j}_{m,n} & \mathbf{C}^{z_j}_{m,n}
    \end{bmatrix} \in \mathbb{R}^{1 \times 3}, \quad m, n \in \{ 1,2,3 \}.
\end{align*}

Then, stacking each element $\mathbf{C}^{x_j}_{m,n}$ as in (\ref{th5}) leads to
\begin{align*}
    \mathbf{F}^{\mathbf{p}_j}_{m,n} &=\left\{\begin{aligned}
     \frac{1}{\lambda_n - \lambda_m} \frac{\partial \mathbf{u}_m^T \mathbf{A} \mathbf{u}_n}{\partial \mathbf{p}_j}, m\neq n \\
    \mathbf{0}  \qquad\qquad, m=n
    \end{aligned}\right.
\end{align*}
where the vector $\mathbf{u}_m$ and $\mathbf{u}_n$ are viewed constant.

By derivations similar method in (\ref{th9}), we can further obtain the specific form of $\mathbf{F}^{\mathbf{p}_j}_{m,n}$, as follows:
\begin{align*}
    \mathbf{F}^{\mathbf{p}_j}_{m,n} &=\left\{\begin{aligned}
     \frac{(\mathbf{p}_j-\bar{\mathbf{p}})^T}{N(\lambda_n-\lambda_m)} (\mathbf{u}_m \mathbf{u}_n^T + \mathbf{u}_n \mathbf{u}_m^T), m\neq n \\
    \mathbf{0}  \qquad\qquad\qquad, m=n
    \end{aligned}\right.
\end{align*}
And hence (\ref{th6}) becomes
\begin{align}\label{partial_u_pj}
    \frac{\partial\mathbf{u}_k}{\partial \mathbf{p}_j} = \mathbf{U} \begin{bmatrix}
    \mathbf{F}^{\mathbf{p}_j}_{1,k} \\
    \mathbf{F}^{\mathbf{p}_j}_{2,k} \\
    \mathbf{F}^{\mathbf{p}_j}_{3,k}
    \end{bmatrix} = \mathbf{UF}^{\mathbf{p}_j}_{k}
\end{align}

With $\frac{\partial \lambda_k}{\partial \mathbf{p}_i}$ in (\ref{th9}) and $ \frac{\partial\mathbf{u}_k}{\partial \mathbf{p}_j}$ in (\ref{partial_u_pj}), we have:

\begin{align}
    \frac{\partial}{\partial\mathbf{p}_j}\Big(\frac{\partial \lambda_k}{\partial \mathbf{p}_i} \Big)
    =& \frac{2}{N}\bigg(\mathbf{u}_k \mathbf{u}_k^T \frac{\partial (\mathbf{p}_i-\bar{\mathbf{p}})}{\partial\mathbf{p}_j} + \mathbf{u}_k(\mathbf{p}_i-\bar{\mathbf{p}})^T\frac{\partial\mathbf{u}_k}{\partial\mathbf{p}_j} \notag\\
    &+\frac{\partial\mathbf{u}_k}{\partial\mathbf{p}_j} \Big(\mathbf{u}_k^T (\mathbf{p}_i-\bar{\mathbf{p}})\Big) \bigg) \notag \\
    =& \frac{2}{N}\bigg(\mathbf{u}_k \mathbf{u}_k^T \frac{\partial (\mathbf{p}_i-\bar{\mathbf{p}})}{\partial\mathbf{p}_j} + \mathbf{u}_k(\mathbf{p}_i-\bar{\mathbf{p}})^T\mathbf{UF}^{\mathbf{p}_j}_{k}\notag\\
    &+\mathbf{UF}^{\mathbf{p}_j}_{k} \Big(\mathbf{u}_k^T(\mathbf{p}_i-\bar{\mathbf{p}})\Big) \bigg) \label{th8}
\end{align}
It should be noted that $\mathbf{u}_k(\mathbf{p}_i-\bar{\mathbf{p}})^T$ is a matrix but $\mathbf{u}_k^T(\mathbf{p}_i-\bar{\mathbf{p}})$ is a scalar. What is more,
\begin{align*}
    \frac{\partial (\mathbf{p}_i-\bar{\mathbf{p}})}{\partial\mathbf{p}_j} = \left\{
    \begin{aligned}
    \frac{N-1}{N}\mathbf{I}, \quad i=j \\
    -\frac{1}{N}\mathbf{I}, \quad i\neq j
    \end{aligned}
    \right.
\end{align*}
Therefore, (\ref{th8}) can be rewritten as (\ref{th7}). $\quad \blacksquare$

\vspace{-0.2cm}
\bibliography{bare_jrnl}







\end{document}